# SiDiTeR: Similarity Discovering Techniques for Robotic Process Automation

Petr Průcha[1][0000-0003-2197-7825] Peter Madzík[1][0000-0002-1655-6500]

[1] Technical University of Liberec, Studentská 1402/2, Liberec, Czechia
`petr.prucha@tul.cz`

**Abstract.** Robotic Process Automation (RPA) has gained widespread adoption in corporate organizations, streamlining work processes while also introducing additional maintenance tasks. Effective governance of RPA can be achieved through the reusability of RPA components. However, refactoring RPA processes poses challenges when dealing with larger development teams, outsourcing, and staff turnover. This research aims to explore the possibility of identifying similarities in RPA processes for refactoring. To address this issue, we have developed Similarity Discovering Techniques for RPA (SiDiTeR). SiDiTeR utilizes source code or process logs from RPA automations to search for similar or identical parts within RPA processes. The techniques introduced are specifically tailored to the RPA domain. We have expanded the potential matches by introducing a dictionary feature which helps identify different activities that produce the same output, and this has led to improved results in the RPA domain. Through our analysis, we have discovered 655 matches across 156 processes, with the longest match spanning 163 occurrences in 15 processes. Process similarity within the RPA domain proves to be a viable solution for mitigating the maintenance burden associated with RPA. This underscores the significance of process similarity in the RPA domain.

## 1 Introduction

Robotic Process Automation is slowly becoming mainstream technology in various corporate organizations. Unfortunately, even though RPA makes work easier in some ways, it can generate additional work, especially during the running of RPA itself [16]. Very often this happens with companies that cross a critical threshold and fall into an RPA maintenance trap [27]. One way to prevent this, according to RPA developers, is to ensure the reusability of RPA components [8, 16, 26]. With a small number of RPA robots and a small number of RPA developers, this can be easily ensured. With a larger number of RPA robots, larger development teams, outsourcing automation to different development teams in different parts of the world and with the turnover of staff, ensuring the reusability of RPA components is very challenging. Making sure that code quality complies with company norms during development is also challenging. For this reason, software developers should refactor their code to be more efficient and serviceable. Hence, it is advisable to refactor the RPA code as well, so that the code components are reusable. As in software development, refactoring can be done backwards.

Aim of the research: *To explore the possibility of finding similarities or identical parts in an RPA process for refactoring if many automations were developed by people who no longer work in a particular company, or if the development was outsourced.*

There is an area in business process management that addresses a similar problem and then tries to find identical processes within an organization, or across manufacturing plants, or after a merger/acquisition. However, these techniques have focused on processes that are not automated. The most commonly used sources for analysis are process logs, natural language content, graph structures, Petri Nets, and BPMN notation [11, 31]. None of these methods are primarily intended for the RPA area. Therefore, input data, which for RPA may be the code of an RPA bot or possibly the log records from RPA bots, are not considered. However, using the foundation of these techniques can help answer our research question and achieve better maintainability by finding parts from RPA code to refactor into reusable components.

The need for a new similarity algorithm comes from the desire to deal with the maintenance trap. The current algorithms and solutions are not compatible with RPA processes or logs. Many current discovery techniques are discussed in the section titled Related Work. While these techniques propose interesting ideas which inspired our solution here, they would be hard to use in the RPA domain or would not be especially effective. Firstly, all currently used algorithms would need a certain amount of data preparation before their application. And then, after all of the transformations, there could arise certain problems related to the specifics of RPA technologies and the structure of process flow. For example, the process inquiry can deviate from reality. The RPA technology sometimes needs to add extra activities to the flow in order to function properly, for example exceptions which account for a loading screen. These extra activities would be problematic because in a standard graphical visualization as a BPMN or a Petri Net, these activities would not be covered. Also, the structure of the RPA code can be more problematic due to the fact that many activities are nested inside other activities. Before the analysis, it is important to flatten the process structure in order to perform an analysis. Lastly, the effectiveness of non-RPA algorithms can be lower, because in a computer environment, it is possible to perform the same action a different way and get an identical output. Our dictionary feature can recognize process activities which are different, even when the activities yield the same output. This extends the pool of similar or same activities. This increases the number of criteria for using algorithms from related work that can be used in the RPA domain after minor or major changes. These criteria will be introduced in the Related Work section.

In this article, we propose that Similarity Discovering Techniques for RPA processes shall be identified as SiDiTeRs. A SiDiTeR is a technique for searching for similar parts of RPA code which could be refactored into reusable components. A SiDiTeR is specially designed for use with UiPath RPA processes, currently the most used RPA tools [38]. The approach can be extended to other commercial RPA solutions in order to discover similarities in RPA processes. Our techniques promise to efficiently discover similar patterns in a sequence of activities to later maximize the ability to leverage the benefit of reusability of the RPA components.

The main contributions of this new algorithm for identifying process similarity in RPA processes are:

- Its ability to work on RPA designs or RPA process logs
- By design it works with the specifics of RPA technologies, like process structure and process flow
- A dictionary feature is provided to extend potential matches and cover identical outputs

In this article we first analyze the previous work related to our approach. Subsequently we describe the use of SiDiTeRs in detail as a method for RPA process similarity discovery. We follow with an evaluation of the method and a conclusion of the work.

## 2 Related work

There are already other approaches for discovering process similarity. Therefore, in this section we will analyze other approaches where a discovery approach is used, what input data is needed, and also how much these approaches comply with our criteria for RPA. We assume that after tuning all of the algorithms, they could at least partly be used in the RPA domain. For example, after converting the RPA processes to another format, a certain approach could be used. For the analysis of other approaches we will classify them based on the publication on process similarity by Schoknecht et al. and Dijkman et al. [6, 29]. Most authors use more than one of these approaches to compare process similarity. Process similarity approaches are:

**Behavioural similarity** methods usually use execution traces of process and then analyze the change in execution states or the behaviour of the flow. That means that they check individual states and their changes.

**Natural language similarity** methods use natural language to try to find similarity in labels of activities. Many other approaches use both syntactic and semantic aspects of language to analyze similarity.

**Graph or structural similarity** methods consider graph structure or business process-aware control flow. Various techniques like the graph edit distance technique or the block structure technique are used to measure the similarity between process models based on their graph structure and control flow.

**Attribute Similarity** methods examine the similarity between the attributes of each activity that are required for the successful execution of that activity in the process.

The criteria for determining if related algorithms (after the necessary changes) have the ability to work effectively in the RPA domain can be summarized from the introduction of this paper. The criteria are:

1. The ability to correctly interpret RPA processes from the RPA design or an RPA log with all of the nested activities inside.
2. The ability to handle the extra activities in the RPA processes that will not be displayed in a graphical visualization of the process.
3. The ability to cover different activities with the same output.

An analysis of the criteria for a match is presented in the last column of Table 1. An analysis of related works for determining which approaches and inputs could be exploited for this study was carried out according to Figure 1. Scopus and Web of Science databases were used to search for related works. All non-BPM records were excluded from the search results, including those from manufacturing, computer science (CPU related), databases, web services, and psychology. We also excluded works related to BPM if they were not relevant for generating similar processes or if the records were not accessible. In this eligibility screening, we also excluded records which did not provide a new method or algorithm for analyzing the process similarity or if they had not yet been validated on any processes.

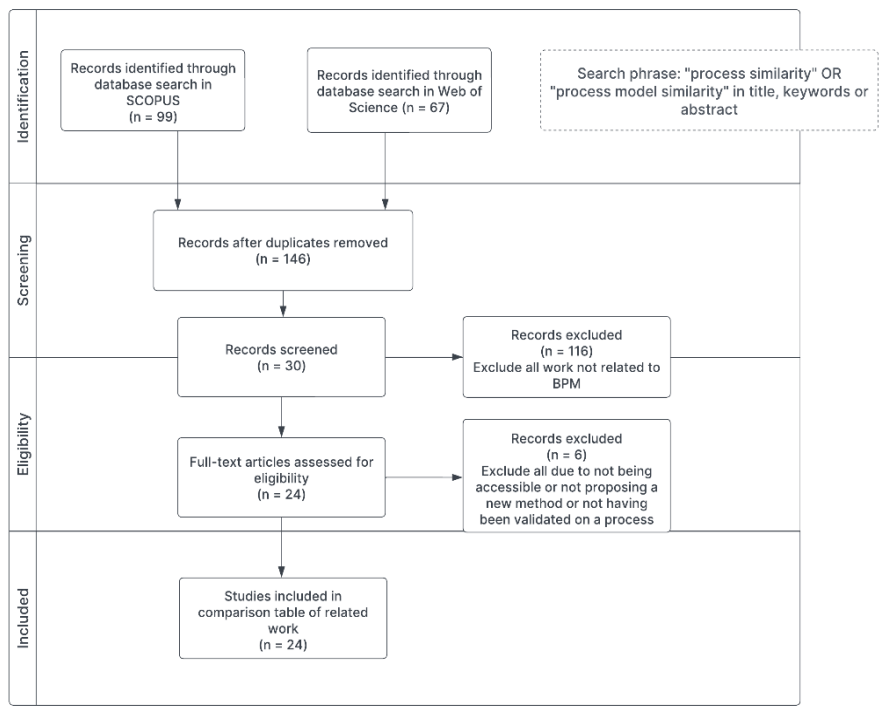

**Fig. 1.** Related work procedure

The result of the analysis of each approach is shown in Table 1. In Table 1 there are only the publications that passed through the filter. Our search phrases are shown in Fig 1. Schoknecht et al. [29] conducted a similar literature review and found 123 rele-vant publications. However, they also used phrases and keywords which were older and, according to them, no longer used today.

**Table 1.** Related work comparison table

| Publication | Type Similarity | Format of input data | Criteria match |
| --- | --- | --- | --- |
| Ye et al. [34] | Graph similarity | Connected Graph | Low |
| Garcia et al. [11] | Graph similarity | (BPMN) 2.0.2 | Medium |
| Pei et al. [25] | Behavioural similarity | Petri Net | Low |
| Niu et al. [24] | Behavioural similarity | Token Logs | Very Low |
| Liu et al. [19] | Behavioural and graph similarity | DWF-nets | Low |
| Sohail et al. [30] | Natural language and behavioural similarity | XML | Medium |
| Zeng et al. [35] | Behavioural similarity | Role relation network | Very Low |
| Zhou et al. [37] | Natural language and behavioural similarity | Business process graph + process log | Very Low |
| Liu et al. [20] | Behavioural similarity | Business process graph | Very Low |
| Valero [32] | Behavioural similarity | Petri Nets | Low |
| Klinkmuller and Weber [18] | Behavioural similarity | Control flow log | Very Low |
| Cao et al. [5] | Graph and behavioural similarity | Petri nets or BPMN | Low |
| Amiri and Koupaee [3] | Structural, attribute behavioural similarity | BPMN | Medium |
| Figueroa et al. [9] | Natural language and structural similarity | Business process in XML | Medium |
| Montani et al. [22] | Structural similarity | Process log | Medium |
| Yan et al. [33] | Attribute similarity | BPMN notation | Medium |
| Niemann et al. [23] | Natural language and graph similarity | SAP reference model | Very Low |
| Dijkman et al. [6] | Behavioural, natural language and graph similarity | SAP reference model | Very Low |
| Zha et al. [36] | Behavioural similarity | Transition adjacency relation set | Very Low |

| Lu et al. [21] | Structural, Behavioural, and natural language | Business process constraint network (BPCN), and process variant repository (PVR) | Low |
| Jung et al. [15] | Structural similarity | Non specified process model is converted to: weighted Complete Dependency Graph (wCDG) | Very Low |
| Dijkman et al. [7] | Natural language and graph similarity | SAP reference model | Low |
| Jung and Bae [14] | Behavioural similarity | Weighted complete dependency graphs, | Very Low |
| Huang et al. [12] | Graph similarity | Weighted complete dependency graphs, | Very Low |

As shown in Table 1, most of the authors used more than one type of similarity techniques. None of the studies focused on RPAs, nor did they utilize RPA source codes or log information. This is confirmed by Schoknecht et al. [29] in their literature review. Most approaches would require transforming the RPA process into a specific input format in order to be usable. For example, converting RPA code into BPMN has already been proposed in some approaches: [10, 13, 28]. The transformation would then be less demanding than with other approaches. The least amount of effort for utilizing an existing method for finding similarity would be to use methods that utilize process logs [22, 37], or other studies that did not appear in the searched results [1, 2].

In Table 1, the criteria match column shows a range of values from very low to very high. These values indicate a match with the criteria presented earlier in this paper. None of the techniques in Table 1 would fulfil all of the criteria. The closest ones were the algorithms which used similar input data to RPAs such as process logs or XML, or which made use of the BPMN format because of its easy transformation from RPA code. Also, some algorithms were valued higher because of a natural language similarity, attribute similarity or other similarity approaches which would be useful in the RPA domain.

## 3 Description of method

Our proposed method SiDiTeR (Similarity Discovering Techniques for RPA) uses natural language-based and graph similarity-based methods. The method is composed of three main parts. The first part is the decomposition of the RPA process/design. The second part of SiDiTeR focuses on natural language matching. The activities from RPA process are compared with activities in a provided dictionary feature (later referred as dictionary **Δ**), and this then produces an abstract (meta) process. The third part of SiDiTeR is the use of the longest common sequence (LCS) algorithm to find the longest sequences in the processes.

### 3.1    SiDiTeR

In the first part, SiDiTeR decomposes the source code of the RPA process, referred to then as the RPA design. From the design, we extract all of the activities with a name $\alpha$. We preserve the order of activities $\alpha$ in the RPA design. Technically, we extract the activity names after the colon tag starting with <ui: from the XAML files. An example is <ui:ReadRange. We extract just the name ReadRange from the text. Thanks to the decomposition, we are able to have an RPA design activity list $\mathbf{A}$ for each process that we decompose this way and save to a list of all activities $\mathbf{A}$.

SiDiTeR then creates a new activity list $\lambda$ for each design. Then it searches through all activities $\alpha$ in the activity list $\mathbf{A}$ and looks for a match in the dictionary of identical activities $\Delta$ (see Table 2). If no match is found, it adds the activity to the new list $\lambda$ with an original name. When a match is found between activity $\alpha$ and activity $\delta$ from the dictionary $\Delta$, activity $\alpha$ is assigned a more abstract description (a meta-action name in Table 2) of activity $\delta$ that describes what the activity does. This results in a more abstract process i.e. meta process of the activity, which is stored in the newly created list $\lambda$. This results in a list of lists denoted as $\Lambda$. This process is visualized for an example in Fig 2.

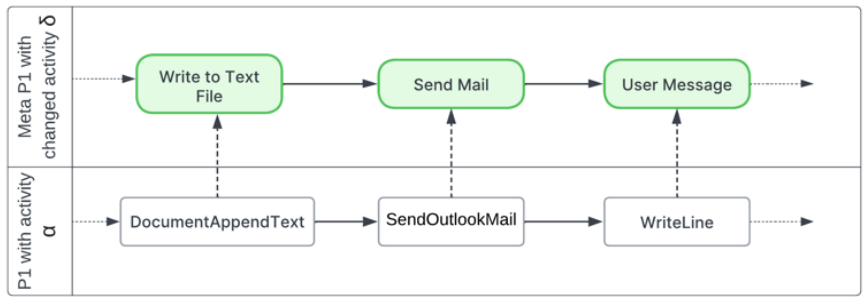

**Fig. 2.** Conversion to a meta process

The third part of SiDiTeR is a search for the longest common sequence for every meta process $\lambda$ saved in $\Lambda$. The longest common sequence algorithm finds identical sequences in all newly made meta processes. The found sequences have to be equal to or longer than 3 activities in order to qualify for saving. The saved activities allow the user to effectively search for similar processes activities which can be then refactored. The user later has to make decisions if the component is the same and should be refactored into reusable components for another RPA process. An example of a found common sequence in two processes is shown in Figure 3. For understanding this process better, a description of the code is written below. See Code 1.

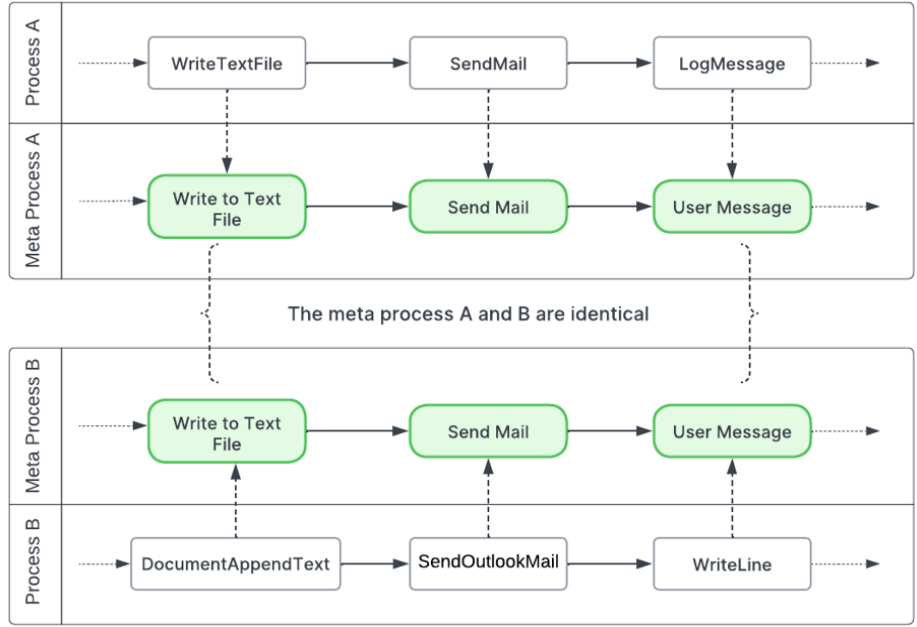

**Fig. 3.** Example of comparing meta processes

**Code 1.** Pseudocode of the SiDiTeR

```
list of activities A
list of lists Λ
for `A in A:
  new list λ
  for α in `A:
    if α match δ in Δ:
      λ add δ
    else:
      λ add α
    Λ add λ
function LongestCommonSequence(Λ):
  return lcs > 3:
```

### 3.2 Dictionary creations for SiDiTeR

The dictionary in Table 2 was created based on activities that were available in Ui-Path Studio, version 2023.4.0-beta.12241 with a community license, with the UiPath packages for OCR, Excel, Word, Ui Automation, Mail and System Activities all installed. The dictionary was created as follows: we tried to find all activities that have the same or similar output but can be achieved by different activities. We only looked for activities that can be interchanged. We were not looking for sequences with the same output. The only exception was for copy-paste activities called SetToClipboard - NkeyboardShortcuts, which also work together as a sequence for writing. Using these actions, SetToClipboard (setting text into the clipboard) and Nkeyboard-Shortcuts (for pasting), are identical to how a user would use copy and paste on a PC, i.e., Ctrl + C and Ctrl + V. In the second column of Table 2, the **δ** activities are grouped by the same meta-action named in the first column. The activity names in the second column come from the UiPath activity names. The same names can be seen in the RPA process source code in the XAML file and also in the UiPath user interface.

**Table 2.** Dictionary **Δ**

| Meta Action Name | Activity Name |
|---|---|
| Write in UI | NTypeInto, SetToClipboard - NKeyboardShortcuts, CVTypeIntoWithDescriptor |
| Write to Text File | WriteTextFile, WordAppendText, DocumentAppendText, AppendLine, DocumentReplaceText, WriteTextFile,NTypeInto |
| Write to Spreadsheet | WriteCSVFile, WriteCellX, AppendCsvFile, WriteRangeX, AutoFillX, ExportExcelToCsvX, InvokeVBAX,CopyPasteRangeX, AppendRangeX, AutoFitX, FindReplaceValueX, AppendRange, WriteCell, WriteRange, ExecuteMacroX, OutputDataTable, AddDataRow, UpdateRowItem, NTypeInto |
| Creation of Data Objects | BuildCollection<Object>, CreateList<Object>, BuildDataTable |
| Write to Data Objects | AppendItemToCollection<Object>, AppendItemToList<Object>, UpdateListItem<Object>, AddDataRow, UpdateRowItem |
| SAP login | Login, Logon, |
| OCR | GoogleCloudOCR, MicrosoftAzureComputerVisionOCR, CjkOCR, GoogleOCR, UiPathDocumentOCR, UiPathScreenOCR |
| Send Mail | SendMail, SendOutlookMail, SendMailX |
| Receive Mail | GetPOP3MailMessages, GetOutlookMailMessages, GetIMAPMailMessages |
| Save Mail | SaveMail, SaveOutlookMailMessage, SaveMailX |
| User Message | LogMessage, WriteLine |
| Get text | CVGetTextWithDescriptor, NGetText, GetOCRText |
| Click | CVClickWithDescriptor, Nclick, ClickOCRText |
| Hover | CVHoverWithDescriptor, Nhover, HoverOCRText |
| Highlight | CVHighlightWithDescirptor, Nhighlight |
| Extract DataTable | CvExtractDataTableWithDescriptor, NExtractData |
| Read File Text | DocumentReadText, WordTextRead, ReadTextFile |
| Save to clipboard | SetToClipboard, CopySelectedText |
| Loop | ForEach<Object>, InterruptibleWhile,InterruptibleDoWhile, ParallelForEach<Int32> |

| Condition | If, IfElseIf, Switch<Int32> |
|-----------|------------------------------|

## 4    Evaluation

Our SiDiTeR approach, as presented in the previous section, was tested on a real RPA process made for UiPath. We programmed SiDiTeR in Python 3.11 to evaluate our approach. The repository with the sample processes is publicly available[1]. We evaluated the effectiveness of SiDiTeR on 156 UiPath process designs. Among the processes were 120 various sample processes, such as setting up an email account, calculator, robotic enterprise framework, executing commands in PowerShell and many others. The processes were in .xaml format and came from public repositories from GitHub or UiPath. In the dataset there were also 36 corporate automations from the banking industry which are not publicly available, and they are under a non-disclosure agreement. The corporate process comes from one banking company, and their process is used in the UiPath Robotic Enterprise Framework for building RPA processes. This is nicely presented in the results, where 15 files from 36 corporate process files have the longest common sequence of 163 same activities in the files. The second longest common sequence is 36 activities, and it comes from a different version of robotic enterprise framework files. The rest of the sequence is much shorter, and it would be important go through the activities manually and evaluate them. All of the results from SiDiTeR are presented in Table 3. In total, we were able to discover 655 matches among the tested xaml files.

**Table 3.** Results

| Length of longest sequence | Number of found values |
|-----------------------------|------------------------|
| 3                           | 481                    |
| 4                           | 125                    |
| 5                           | 30                     |
| 6                           | 2                      |
| 9                           | 1                      |
| 36                          | 1                      |
| 163                         | 15                     |

At the outset, we proposed the following research aim:

*To explore the possibility of finding similarities or identical parts in an RPA process for refactoring if many automations were developed by people who no longer work in a particular company, or if the development was outsourced.*

---

This research paper demonstrates that it is possible to identify similar or identical parts in an RPA process. The results show that SiDiTeR can identify the same or similar activities across RPA processes and help the RPA developers or RPA maintenance team identify the activities which are candidates for refactoring.

## 5 Discussion and Limitations

We have proposed a new method for discovering similarity in RPA processes (SiDiTeR). SiDiTeR uses an RPA design for the analysis of similar parts of different processes. This helps to refactor RPA code into reusable components more easily. The results show that SiDiTeR is able to find candidates among RPA processes for refactoring. As mentioned in the introduction, this is one of the solutions for overcoming an RPA maintenance trap, as the whole portfolio of RPA bots will then be more easily governed [16]. To find out which part of an RPA process should be refactored into components, process similarity techniques can be used.

In the field of process similarity, there has been a decline in the number of new works published [29]. The use of process equivalence and process similarity techniques in the field of RPA can be a new spark for more research and publications in the field. With a higher number of RPA automations, there will be a higher demand for making the automations sustainable and avoiding the RPA maintenance trap. As seen from the related works, no technique has addressed this topic yet. Thus, this could be an impulse for using process similarity in another practical application.

We are aware of certain limitations that our approach currently has. One of the concerns is that SiDiTeR works only with UiPath designs, and the dictionary is made for UiPath activities. This limitation concerning UiPath designs is easily addressable, at least partially, and it would be enough to decompose the activity names from the source code of another platform. The limitation concerning the dictionary is more complicated, as partial knowledge of the platform is needed to create a similar dictionary. It is likely that the size of the dictionary will be different for different platforms. In certain cases, such as writing vs copying and pasting text, these activities can be adopted one to one for other platforms. When creating the dictionary for our study, only activities that had identical or similar resulting actions were used. The dictionary could be extended to include sequences where the output of the activities is also identical, but the result achieved is made up of multiple actions such as: clicking in the UI vs using a keyboard shortcut; or, for example, using the UI instead of using the API. Experienced programmers are likely to use the most efficient path, but for junior development or citizen development, inefficient sequences are likely to occur [17, 27].

SiDiTeR can also raise questions about why we use process similarity techniques for processes instead of techniques from the computer science field, even though RPA is software. This is a justified question because there are already techniques for code refactoring. For example, a systematic literature review from 2020 [4], analyzed 41 techniques concerning automatic software refactoring. But we focused more on pro-cess similarity due to the fact that RPA process (code) can also be analyzed as a process. RPA as a process is more understandable to non-technical users, citizen developers and process owners. The understanding by stakeholders of a process can by crucial for the additional validation of refactoring of the correct part of a process. The main advantage of SiDiTeR techniques is that they can be used on the source code of RPA or also on the process log to analyze the RPA as a process.

Another limitation may be the accuracy of SiDiTeR, where in some cases the activities are not identical but will still be included, even though they are different processes i.e. false positives. Accuracy could be increased by using parameters and incorporating attribute similarity into SiDiTeR. This approach would then be even more efficient for users who will evaluate the results. There is an opportunity for extending this research further, for the purpose of identifying the right candidates for refactoring among RPA processes more precisely.

## 6　　Conclusion

Finding similarity in the RPA domain is very useful, because it can be used for refactoring. The refactoring of RPA processes will be one of the crucial components for future RPA governance, since the same parts of RPA code can be refactored into components and shared across a portfolio of RPA bots. We have presented a new approach for detecting identical or similar parts in RPA processes called SiDiTeR. SiDiTeR is designed with RPAs in mind, and can easily read RPA code or process logs with nested activities and handle extra activities in processes. It can also deal with different activities with the same output, which is crucial for complex refactoring. Our approach was tested on 156 RPA processes. The longest match we discovered was with 163 activities across 15 processes and 655 matches among RPA processes. These results challenge future researchers to find ways to identify parts of RPA which could be more precise, and thus allow for a more convenient search method for suitable components for refactorization.

**Acknowledgment:** This research was made possible thanks to the Technical University of Liberec and the SGS grant number: SGS-2023-1328. This research was conducted with the help of Pointee.